\documentclass[letterpaper, 10 pt, conference]{ieeeconf}  
\pdfoutput=1

\IEEEoverridecommandlockouts                              

\usepackage[skip=0pt,font=small,labelfont=bf]{caption} 
\usepackage{booktabs}
\usepackage{hyperref}
\usepackage{graphicx}
\usepackage{float}
\usepackage{amsmath, amssymb}
\usepackage{amsfonts}
\usepackage{bm}
\usepackage{multirow}
\newcommand{\x}{x} 
\newcommand{\ov}{\bm{o}} 
\newcommand{\s}{s} 
\newcommand{\fsdf}{f_{SDF}} 
\newcommand{\qv}{\bm{q}} 
\newcommand{\fk}{\texttt{FK}} 
\graphicspath{ {./images/} }
\newcommand\numberthis{\addtocounter{equation}{1}\tag{\theequation}}

\usepackage[normalem]{ulem}

\usepackage{float}
\usepackage{subfig}
\usepackage{epsfig}
\usepackage{xcolor,import}
\usepackage{transparent}
\usepackage{upgreek}
\usepackage{tikz}
\usetikzlibrary{shapes,calc,positioning,arrows,}
\pgfarrowsdeclarecombine{twotriang}{twotriang}
{stealth'}{stealth'}{stealth'}{stealth'} 
\usepackage{cuted}
\usepackage{adjustbox}
\usepackage{pgfplots}
\pgfplotsset{major grid style={dotted,green!50!black}}
\usetikzlibrary{patterns}

\usepackage{colortbl}

\let\labelindent\relax
\usepackage{enumitem}
\usepackage{wrapfig}

\DeclareMathOperator*{\argmin}{argmin}
\usepackage{cite} 

\usepackage{fixme}
\fxsetup{status=draft, theme=color}
\definecolor{fxtarget}{rgb}{0.8000,0.0000,0.0000}
\definecolor{fxnote}{rgb}{0.0000,0.0000,0.8000}

\newcommand{\shrinka}{\def\baselinestretch{0.993}\large\normalsize}

\title{\LARGE 
  Learning Continuous 3D Reconstructions for Geometrically Aware Grasping
}

\author{
  Mark Van der Merwe$^{1}$, Qingkai Lu$^{1}$, Balakumar Sundaralingam$^{1}$, Martin Matak$^{1}$, and Tucker Hermans$^{1,2}$%
  \thanks{$^{1}$School of Computing and Robotics Center, University of Utah, Salt Lake City, UT 84112, USA.}
\thanks{$^{2}$NVIDIA Research, Seattle, WA 98105, USA}%
}

\definecolor{lightgreen}{RGB}{240,249,232}
\definecolor{darkgreen}{RGB}{186,228,188}

\definecolor{lightblue}{RGB}{123,204,196}
\definecolor{darkblue}{RGB}{43,140,190}

\begin{document}
\setcounter{figure}{1}
\makeatletter
\let\@oldmaketitle\@maketitle
\renewcommand{\@maketitle}{\@oldmaketitle
\begin{center}
  \centering     
  \includegraphics[width=0.9\textwidth] {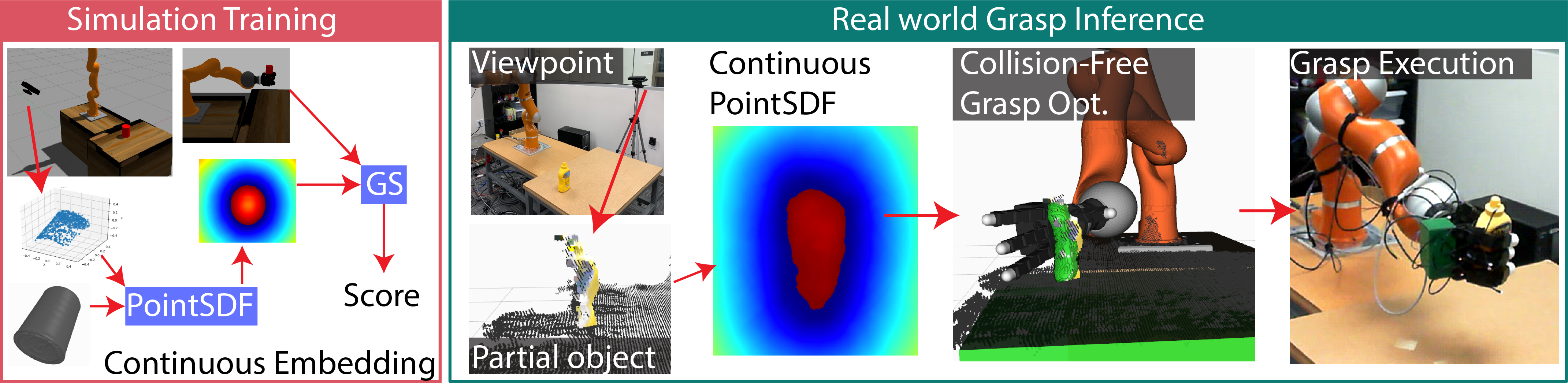}
\end{center}
  \footnotesize{\textbf{Fig.~\thefigure:\label{fig:title}}~ We train a continuous signed distance function embedding~(PointSDF) for the partial object pointcloud, and a grasp success~(GS) model in simulation. Leveraging these simulation learned models in a gradient-based grasp optimization, we enable collision-free grasping of novel objects in the real world.}\vspace{-12pt}
  \medskip}
\makeatother
\shrinka
\maketitle
\shrinka
\thispagestyle{empty}
\pagestyle{empty}


\begin{abstract}
Deep learning has enabled remarkable improvements in grasp synthesis for previously unseen objects from partial object views. However, existing approaches lack the ability to explicitly reason about the full 3D geometry of the object when selecting a grasp, relying on indirect geometric reasoning derived when learning grasp success networks. This abandons explicit geometric reasoning, such as avoiding undesired robot object collisions. We propose to utilize a novel, learned 3D reconstruction to enable geometric awareness in a grasping system. We leverage the structure of the reconstruction network to learn a grasp success classifier which serves as the objective function for a continuous grasp optimization. We additionally explicitly constrain the optimization to avoid undesired contact, directly using the reconstruction. We examine the role of geometry in grasping both in the training of grasp metrics and through 96 robot grasping trials. Our results can be found on https://sites.google.com/view/reconstruction-grasp/.
\end{abstract}


\section{Introduction}
\label{sec:intro}
The ability to reliably grasp previously unseen objects in multiple environments remains an open challenge in robotics~\cite{Sahbani2012,Bohg2014}. The effects of noisy and partial sensor inputs coupled with the unknown object properties complicate effective grasp synthesis. In this paper, we consider grasping unseen, isolated objects on tabletop environments with multi-fingered dexterous hands. 

While analytical robotic grasp synthesis methods can provide desirable guarantees about grasp performance, they rely on metrics that have failed to generalize to the real world and can fail to perform given perceptual uncertainty; as such, much recent work in robotic grasp synthesis has turned to deep-learning based approaches~\cite{Bohg2014}. Most existing deep-learning approaches are trained in an end-to-end fashion~\cite{Zeng2018,Pinto2016, Levine2018, Mahler2019, Varley2015, Lu2017, Lu2019, Kappler2015, Zhou2017, Liu2019, Veres2017}. That is, the system takes in sensor input, such as an RGB or RGBD image, and outputs a grasp, either via direct regression~\cite{Liu2019, Veres2017}, sampling candidate grasps or motions~\cite{Mahler2019, Levine2018}, or solving an optimization problem leveraging the learned network, either in a discrete ~\cite{Zeng2018, Pinto2016} or continuous~\cite{Lu2017, Lu2019, Zhou2017, Varley2015,liu-arxiv2019-grasp-opt} fashion. As such, there is typically no explicit modeling of the geometry in the scene. Rather, researchers assume the classifier will indirectly learn a geometric understanding of the scene, such that the network will prefer stable grasps that are out of collision. This abandons explicit a priori geometric reasoning, yielding undesirable robot-object collisions~\cite{Zeng2018,Pinto2016, Levine2018, Mahler2019}. We seek to decrease the chance of such collisions by explicitly modeling the 3D environment as part of the grasp planning problem.

The more difficult problem of multi-fingered grasping has similarly followed an end-to-end grasp learning framework \cite{Varley2015, Lu2017, Lu2019, Kappler2015, Zhou2017, Liu2019}. One competitive approach to multi-fingered grasping, achieving state-of-the-art results, relies on performing a continuous optimization over the hand configuration to maximize the likelihood of grasp success~\cite{Lu2017,Lu2019,Zhou2017}. 
In~\cite{Lu2017,lu2020multi}, the continuous grasp optimization is shown to generally achieve higher grasp success rates compared sampling-based~\cite{lenz2015deep,levine2016learning} and regression~\cite{Liu2019,Veres2017} approaches used for grasping with neural networks.
However, similar to other approaches, these optimization-based inference procedures have no explicit understanding of the geometry of the scene and thus may find a solution which causes the hand to be in collision with the environment or even intersect with the object to be grasped. This forces relying on full state knowledge of the environment~\cite{Zhou2017} or abandoning grasp attempts when motion planners fail to find a collision free path to the desired grasp.

The primary obstacle to explicitly incorporating geometric information into grasp synthesis is that these systems have only a single view of the world and thus can only partly understand the object geometry. One approach adopted in the computer vision community is to learn to predict the underlying 3D shape generating the partial view~\cite{Brock2016,Choy2016,Mescheder2019,Park2019,Chen2019}. The recent dominant approach has been to learn a voxel-based object reconstruction \cite{Choy2016} from the partial view; these reconstructions have been utilized in analytical~\cite{Varley2017} and learning-based~\cite{Yan2018a} grasp synthesis systems. Recent work has shown that neural networks can effectively learn implicit shape representations, such as signed distance functions (SDF) or continuous occupancy maps, yielding state-of-the-art 3D reconstruction performance~\cite{Mescheder2019, Park2019, Chen2019}. Learning signed distance function reconstructions yields many desirable improvements to voxel based approaches, including arbitrarily high resolution and mesh-free geometric understanding. Indeed, roboticists regularly use signed distance functions to encode collision constraints in trajectory optimization for motion planning~\cite{Zucker2013}.

We propose utilizing 3D object reconstruction to enable \textit{geometrically-aware grasp synthesis} in a continuous optimization framework~\cite{Lu2017,Lu2019}. At the core of our approach lies a novel implicit surface reconstruction algorithm, \emph{PointSDF}, which directly regresses signed distance functions from point clouds, providing geometrically rich input and output. We leverage these reconstructions to make our grasping system geometrically aware both implicitly and explicitly.

We implicitly encode geometry by introducing the point cloud embedding from PointSDF into a grasp success prediction network~\cite{Lu2017}. We enable explicit geometric reasoning by constraining the optimization of our grasp success prediction network to be collision free. We achieve this by extending previous approaches to learning-based grasp optimization~\cite{Lu2019,Lu2017,Zhou2017} to include the full robot arm configuration, instead of only a 6DOF wrist pose, and add SDF collision constraints between the reconstructed object and all links of the robot. By formulating the optimization in the robot joint space, we ensure not only kinematic feasibility of all synthesized grasps but also Euclidean updates of the gradients in the optimization by propagating learned gradients through the kinematic Jacobian. We examine the efficacy of our approach through real robot grasping experiments on a KUKA LBR4 robot with an Allegro multi-fingered hand.

To summarize, our primary contributions are listed below,

\begin{itemize}[leftmargin=10pt,labelindent=10pt,topsep=0pt]
\item \textbf{3D Reconstruction:} a novel single-view reconstruction learning architecture, PointSDF, that learns a signed distance function implicit surface for a partially viewed object.
\item \textbf{Grasp Success Prediction:} a novel grasp success prediction learning architecture, that implicitly learns geometrically aware point cloud encodings.
\item \textbf{Grasp Synthesis:} an extended formulation of learning-based grasp synthesis as a constrained optimization  problem in the full robot configuration space, ensuring kinematic feasibility and explicit collision avoidance via our learned continuous signed distance function~(PointSDF).
\end{itemize}

We illustrate our key contributions in Fig.~1. We organize the remainder of the paper as follows. In Sec~\ref{sec:3d-recon} we present PointSDF for geometric reconstruction. In Sec.~\ref{sec:grasp-success} we apply PointSDF to grasp success prediction and define the full grasp synthesis optimization in Sec.~\ref{sec:grasp-synthesis}. We discuss the implementations, experiments, and implications of our results in Sec.~\ref{sec:experiments}. We then briefly conclude and discuss directions for future work in Sec.~\ref{sec:conclusion}.


\section{3D Reconstruction via Learned Signed Distance Function}
\label{sec:3d-recon}

\begin{figure*}[]
  \centering
  \includegraphics[width=0.85\textwidth]{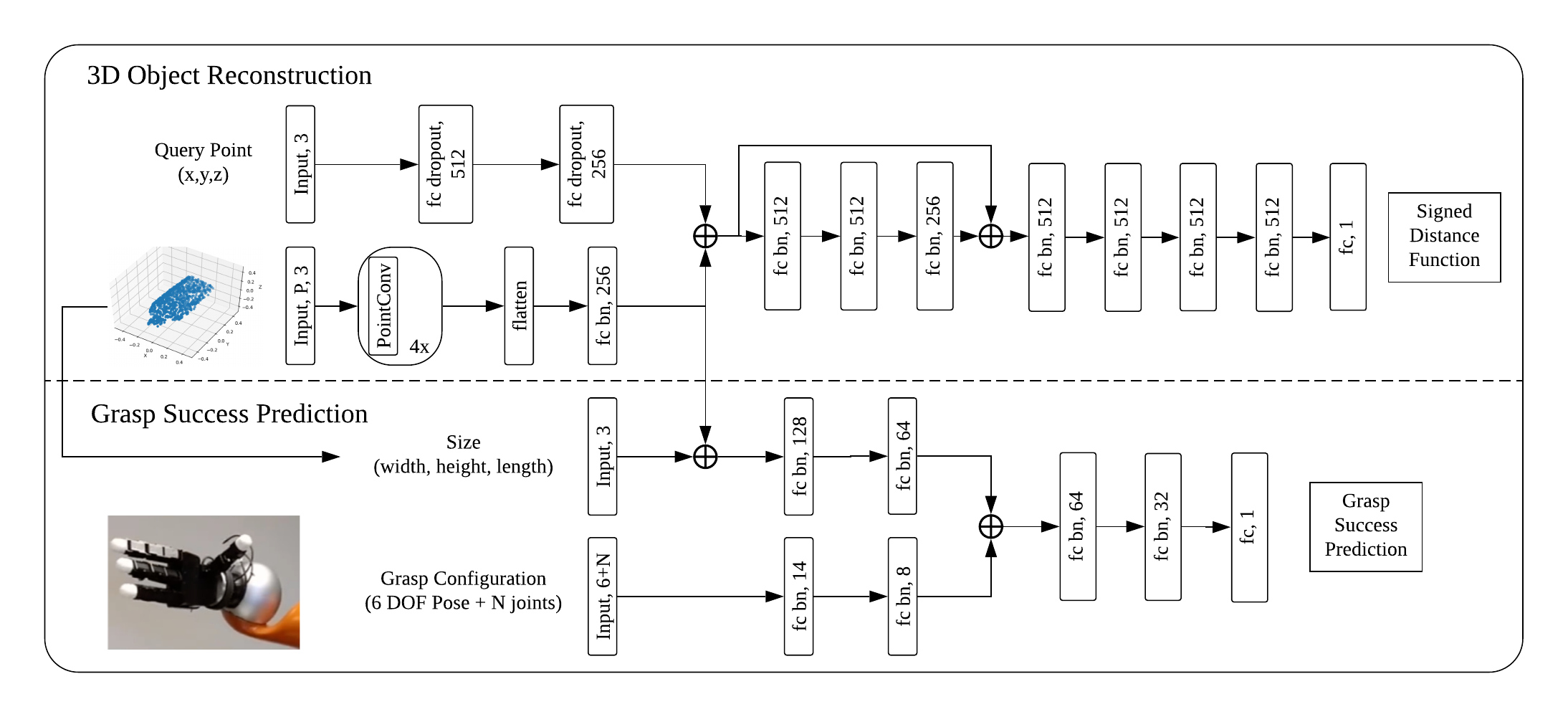}
  \caption{The top network is a 3D reconstruction network that takes in a point cloud and a query point and regresses to signed distance function values for each query point to the surface of the reconstructed object. The bottom network utilizes the point cloud embedding subnetwork, as well as grasp configuration and point cloud size information to predict whether the given grasp configuration will succeed.}
  \label{fig:networks-design}
\end{figure*}

We present a new architecture for predicting a 3D reconstruction of an object from a single view point cloud. Motivated for its use in grasp planning, we desire that our reconstruction approach seamlessly handles seen and unseen objects alike from arbitrary viewpoints and accurately encodes geometric concepts, while efficiently performing inference in terms of both time and space. As such, we propose learning to directly predict the signed distance function, which implicitly represents the object surface as the zero level set of the function. The signed distance function defines the shortest distance between a query point in 3D space and the surface of the object, where distances are negative for points inside the object and positive when outside. We call our architecture PointSDF. Unlike previous iterations of similar design~\cite{Park2019}, we enable single-pass evaluation using a simple encoder-decoder structure.

Given a point cloud view of the object, \(\ov\),  and a query point in 3D space, \(\x\), the PointSDF function, \(\fsdf\) predicts the continuous-valued signed distance from that point to the surface of the fully reconstructed object:
\begin{equation}
  \fsdf(\ov, \x; \theta)=\s;\ \ov\in\mathbb{R}^{P\times 3}, \x \in\mathbb{R}^{3}, \s \in\mathbb{R}
\end{equation}
where $\theta$ represents the parameters of the network. We train our network using the standard mean-squared error loss for regression between the distance predictions from the network and the true SDF values for query points relative to the training objects.

Directly regressing SDF values from the point cloud holds several key advantages that make our PointSDF representation advantageous for robotics applications, especially grasp planning: first, PointSDF can be evaluated at arbitrary resolution without transforming the prediction into a mesh, maintaining accuracy as compared to discretization inherent in earlier voxel-based approaches in robotics~\cite{Varley2017, Lundell2019, Yan2018a}. Nevertheless, a mesh can be extracted if desired by determining the zero isosurface of the SDF, which we can compute via sampling the network throughout the space as shown by~\cite{Park2019, Mescheder2019}.

Another benefit of using a SDF representation via a neural network is that the network implicitly learns geometric gradients. Given a query point $\x$ and observation $\ov$, we can derive a vector pointing towards or away from the true reconstructed object by finding the gradient at the point $\x$, $\partial \fsdf(\ov, \x;\theta) / \partial \x$. We can efficiently compute such gradients using the backpropagation algorithm.

Our architecture builds on recent work deriving convolution operations directly on point clouds \cite{Qi2017,Wu2018}. Unlike typical convolutions which require well structured inputs, these approaches work directly on unstructured point clouds. We embed the point cloud using four PointConv layers~\cite{Wu2018}. This embedding, along with the query point, is passed through several fully-connected layers, leading to a single prediction passed through a tanh activation to get an SDF estimate between -1 and 1. The top half of Figure~\ref{fig:networks-design} shows our network design.

\section{Learning Grasp Success Prediction via 3D Reconstruction}
\label{sec:grasp-success}

Our primary goal is to synthesize high quality grasps for partially viewed objects, which we perform as a continuous optimization over the robot's arm-hand joint configuration. Here we present our design of a learned grasp success metric, that serves as the objective to maximize during grasp synthesis. Following recent, high-performing approaches to multi-fingered grasp planning~\cite{Lu2017,Lu2019}, we model this planning problem as probabilistic inference.

We seek to maximize the posterior probability of a robot arm-hand joint configuration, $\qv$, generating a successful grasp (i.e., $Y=1$) given a point cloud observation, $\ov$, of the target object. For our objective we replace the full configuration, \(\qv = [\qv_h, \qv_a]\), with \(\qv_g=[\qv_h,\fk_p(\qv_a)]\). Here \(\qv_h\) represents the $N$ joint positions of the hand, while \(\fk_p(\qv_a)\) computes 6-DOF palm pose of the robot hand as a function of the robot arm joint state \(\qv_a\). Formally we have the following objective:
\begin{align}
  p(\qv | Y=1, \ov) &\propto p(Y=1|\qv, \ov; \phi)\cdot p(\qv; \psi)\\
                    &\propto h(\qv, \ov; \phi) \cdot g(\qv; \psi) \label{eq:parameterized-prob}
\end{align}
where $\phi$ and $\psi$ parameterize each respective probability distribution. Following Lu et al.~\cite{Lu2017}, we parameterize the grasp success probability distribution \(h(\cdot)\) as a neural network and the grasp prior \(g(\cdot)\) as a Gaussian mixture model (with 2 components) fit to all grasp configurations seen during training.

In order to make our grasp success prediction network geometrically aware, we utilize the same point cloud encoder architecture used in PointSDF to embed the point cloud of the target object. Along with the grasp configuration, the embedding is passed through several fully-connected layers, leading to a single prediction passed through a sigmoid activation to get our grasp success probability estimate. Our network design is shown in the bottom half of Figure~\ref{fig:networks-design}.

\section{Reconstruction-Aware Grasp Synthesis via Constrained Optimization}
\label{sec:grasp-synthesis}
Given the robot joint configuration~$\qv$ encoding the arm and hand joint vectors respectively, we define the grasp synthesis problem as finding a grasp preshape joint configuration for the arm and the hand that enables a collision-free grasp on the object that maximizes the probability of grasp success. This amounts to solving the following constrained, non-convex optimization problem,
\begin{align*}
    \argmin_{\qv} \quad & \max(\beta - s,0)^2 \numberthis \label{eq:f_grasp_score}\\
  \text{s.t.}   \quad & \qv_{min}\preceq \qv \preceq \qv_{max}\numberthis \label{eq:const_jp} \\
                                      & a_{SDF}(M_e,\fk_l(\qv)) > 0 \quad \forall l \in L; e \in E \numberthis \label{eq:const_collision}\\
& f_{SDF}(\ov, \fk_l(\qv)) > 0 \quad \forall l \in L \numberthis \label{eq:const_learned_sdf}
\end{align*}
where $s=-\log(h(\qv_g,\ov))-\alpha \log(g(\qv_g)$ is the cost term derived from the learned grasp success prediction network and prior function. In Eq.~\ref{eq:f_grasp_score}, we turn Eq.~\ref{eq:parameterized-prob} into a least-squares cost to be minimized. As we are combining the log likelihood of a discrete and continuous probability distribution, the range of our objective function in Eq.~\ref{eq:f_grasp_score} is unbounded below; as such, we empirically select some value $\beta$ as a sufficient minimum and square the difference; for our experiments, we use $\beta=-2$. We can derive objective function gradients via backpropagation through our grasp success network.

We split our collision constraints into two parts. First is a known geometry collision constraint where \(M_e\) defines the mesh relating to component \(e\) of the environment (e.g., avoid collision with known table geometry). This utilizes an analytical signed distance function constraint in Eq.~\ref{eq:const_collision}. Second is the unknown target object collision constraint, whose geometry we assume is unknown beyond the single view recieved from a depth sensor $\ov$. This utilizes our reconstruction network $f_{sdf}(\cdot)$ to enforce the constraint in Eq.~\ref{eq:const_learned_sdf}. Because this network takes in the query point and assigns an SDF, we avoid meshing and can directly utilize the learned network to get SDF values for our robot. We compute the query points to check for robot-object collision by computing the forward kinematics of each link for the current robot joint configuration $\qv$ and using each vertex from the link mesh. Each link is assigned the minimum SDF value of all its vertices. As described in Sec.~\ref{sec:3d-recon}, we can derive SDF gradients to move a link out of collision directly through the network via backpropagation. The constraints in Eq.~\ref{eq:const_jp} encode the kinematic limits of the robot joints. 

This formulation allows direct enforcement of geometric constraints that avoids unintentional robot-object contact for the full robot arm. The formulation also ensures all solutions are kinematically feasible, due to the optimization over the full arm-hand joint configuration $q$. 

\section{Experiments}
\label{sec:experiments}

Our experiments seek to examine the efficacy of our reconstruction-based grasping formulation. Specifically, we seek to answer the following questions: 1. How well does PointSDF perform at single depth view reconstruction? 2. How well does our grasp metric perform at predicting grasp success and how does implicit geometric reasoning affect this performance? 3. How well does our grasp formulation work when deployed on a real robot? See videos of our experiments at our website: ~\url{https://sites.google.com/view/reconstruction-grasp/}.

\subsection{PointSDF Reconstruction}\label{sec:recon-results}

We first describe the training procedure for the PointSDF architecture introduced in Sec.~\ref{sec:3d-recon}, then evaluate its performance.

To train PointSDF, we synthetically render, via a simulated camera, the 590 meshes from the Grasp Database~\cite{Kappler2015} and 76 meshes from the YCB Database~\cite{Calli2015} at 200 random orientations each, adding noise to the depth images to reflect Kinect noise. We backproject the rendered points into a 3D point cloud, which becomes the input to our network. We generate SDF query label pairs by rotating the true mesh to the same random rotation for each view and sampling and labeling points from the surface and space around the mesh.

We employ a simple object frame centered in the point cloud that uses the same orientations as the camera frame. By keeping the camera frame orientation and by applying random rotations during data collection, PointSDF is camera-pose invariant. To simplify learning, we also scale all point clouds to fit in a \(1m \times 1m \times 1m\) bounding box. At inference time, SDF estimates can easily be scaled back for use.

To demonstrate how our reconstructed meshes compare to the reconstructions currently used in grasping~\cite{Varley2017,Yan2018a}, we reimplement a representative voxel-based reconstruction model based on a 3D CNN encoder-decoder structure \cite{Brock2016}, which we hereafter refer to as VoxelCNN. We quantify 3D reconstruction performance with three metrics computed between the reconstructed test object and the ground truth mesh: the volumetric IoU, Chamfer-$L_1$ distance, and a normal consistency score (for description of metrics, we refer the reader to~\cite{Mescheder2019}). To recover the mesh from our implicit surface model, we re-implement a hierarchical sampling reconstruction method from ~\cite{Mescheder2019}, and sample to a resolution of $512^3$.

\begin{table}
  \centering
  \caption{Reconstruction performance on different metrics (standard deviation in parentheses).}
\label{tab:reconstruction_quantitative}
\begin{tabular}{ l l c c }
  \toprule
  
  \multirow{2}{*}{\bf Metric} & \multirow{2}{*}{\bf Method} & \multicolumn{2}{c}{\bf Dataset} \\ \cline{3-4}
    & &  Grasp Database &  YCB \\
  \toprule
  \multirow{2}{*}{IoU} & PointSDF & \textbf{0.71933 (0.18807)} & \textbf{0.56975 (0.28807)} \\
  & VoxelCNN & 0.53172 (0.20040) & 0.44039 (0.23304) \\
  \midrule
  \multirow{2}{*}{Chamfer-$L_1$} & PointSDF & \textbf{0.00150 (0.00432)} & \textbf{0.00256 (0.00237)} \\
  & VoxelCNN & 0.00296 (0.00116) & 0.00458 (0.00230) \\
  \midrule
  Normal & PointSDF & 0.83611 (0.06540) & 0.77919 (0.10078) \\
  Consistency & VoxelCNN & \textbf{0.83625 (0.06094)} & \textbf{0.78665 (0.07273)} \\
  \bottomrule
\end{tabular}
\end{table}

\begin{figure}[]
  \centering
  \includegraphics[width=0.98\columnwidth]{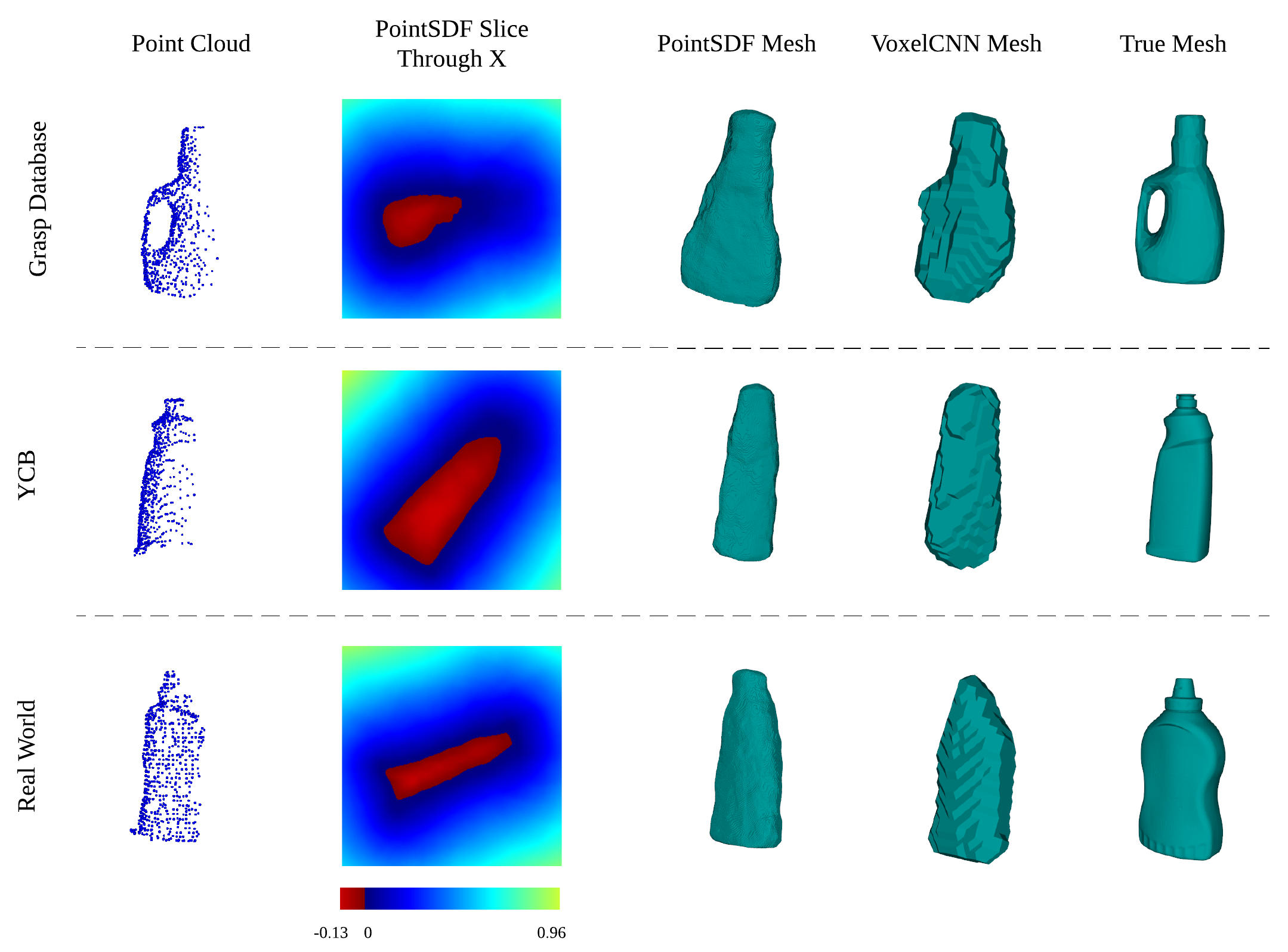}
  \caption{Qualitative comparison of reconstruction results with objects from the Grasp and YCB dataset, as well as the YCB mustard with a point cloud from the real camera. We also show samples in a plane through the target object.}
  \label{fig:recon-qual-comparison}
\end{figure}

In Table~\ref{tab:reconstruction_quantitative}, we see that PointSDF outperforms VoxelCNN on the IoU and Chamfer-$L_1$ metrics and nearly matches VoxelCNN on normal consistency, indicating increased geometric understanding. In Figure~\ref{fig:recon-qual-comparison}, we show representative object reconstructions on previously unseen objects from both datasets. We see that PointSDF reconstructions are smoother and retain finer details as compared to VoxelCNN results. The PointSDF slices also show a desirable gradient in predictions with a clear zero level set.

\subsection{Grasp Success Network}\label{sec:grasp-success-results}

We now describe our specific grasp configuration $\qv_g$, our data collection and training procedure, and evaluate our grasp success network performance.

We conduct all training and experiments using the four-fingered, 16 DOF Allegro hand mounted on a Kuka LBR4 7 DOF arm. As described in Sec.~\ref{sec:grasp-success}, we use the palm pose of the grasp (in the object frame used for reconstruction) and $N=8$ joint values, representing the two joints of each finger closest to the palm.

We collected simulated grasp data for grasp model training using our robot hand-arm setup inside the Gazebo simulator with the DART physics engine\footnote{\url{https://dartsim.github.io/}}. We use the built-in Gazebo Kinect camera to generate point clouds. We collected training data using a heuristic, geometry-based grasp planner~\cite{lu2020multi} and randomly sample joint angles for the first two joints of all fingers within a reasonable range, fixing the last two joints of each finger to be zero. More details of our grasping data collection can be seen from~\cite{lu2020multi}. In total, we train with 7290 grasp examples, and test on a leave-out set of 1821 grasps.

To determine how implicit geometric reasoning effects the performance of grasp success prediction, we train two variations of our network from Sec.\ref{sec:grasp-success}: first, we train the point cloud embedding from scratch, only with regards to the grasp dataset (referred to as ``PointSDF-Scratch''). Second, we train the point cloud embedding on reconstruction, then lock the embedding network and only update the grasp metric model (referred to as ``PointSDF-Fixed''). The former acts as a baseline which learns geometric reasoning implicitly from the downstream task, whereas the latter seeks to impose geometric reasoning by first training the embedding on the reconstruction tasks.

To compare to voxel-based approaches, we replace the PointSDF embedding with a voxel-based embedding from Sec.~\ref{sec:recon-results}, and setup the same two variations (``VoxelCNN-Scratch'' and ``VoxelCNN-Fixed'')

\begin{figure}
  \centering
   \begin{adjustbox}{minipage={0.8\columnwidth}}
    \begin{tikzpicture}
      \begin{axis}[
          ybar, ymin=0.0, ymax=1.0,
          nodes near coords,
          ylabel={Grasp Success F1-Score},
          axis lines*=left,
          y label style={at={(axis description cs:0.045,0.5)},anchor=north},
          symbolic x coords={VoxelCNN-Scratch, VoxelCNN-Fixed, PointSDF-Scratch, PointSDF-Fixed},
          xtick=data,
          x tick label style={font=\scriptsize,rotate=0,yshift=0.0cm,xshift=0.0cm,text width=0.8cm,align=center},
          xticklabels={VoxelCNN-Scratch, VoxelCNN-Fixed, PointSDF-Scratch, PointSDF-Fixed},
          ymajorgrids=true,
          ticklabel style = {font=\scriptsize},
          legend style={font=\scriptsize, draw=none, fill=none},
          ylabel style = {font=\scriptsize},
          bar width = 20pt, height=3.5 cm, width=1.05\columnwidth, 
          legend style={area legend, at={(1,1.25)}, anchor=north east, legend columns=2, },
          legend image code/.code={%
            \draw[#1] (0cm,-0.1cm) rectangle (2mm,1mm);},
        ]
        \addplot [fill=darkblue] coordinates {
          (VoxelCNN-Scratch, 0.4) (VoxelCNN-Fixed, 0.53614) (PointSDF-Scratch, 0.63992) (PointSDF-Fixed, 0.58072)};
      \end{axis}
    \end{tikzpicture}
  \end{adjustbox}
   \caption{Grasp success prediction performance across different embedding approaches.}
     \label{fig:grasp_success_quantitative}
\end{figure}
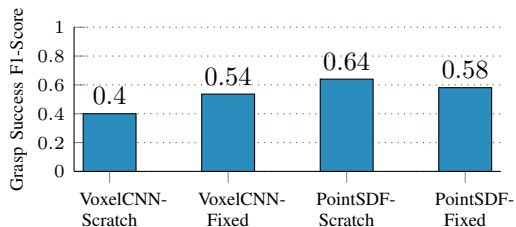

We show in Fig.~\ref{fig:grasp_success_quantitative} the F1-score of each classifier (with threshold 0.5 on prediction) against the test set from our simulated grasps. We highlight two key observations from our results. First, we notice that both networks based on the point cloud embedding~\cite{Wu2018} outperformed the voxel-based networks. Second, locking the embedding network based on the reconstruction slightly decreased the performance of the classifier when using the PointSDF encoder. We speculate that this could be due to, a) the structure of PointConv layers in the encoder, and b) the small size of our training set. We notice that for the VoxelCNN method this role was reversed, indicating that the CNN structure appears to encourage overfitting.

\subsection{Grasp Synthesis}\label{sec:grasp-synthesis-results}
\begin{figure}
  \centering
  \includegraphics[width=0.6\columnwidth]{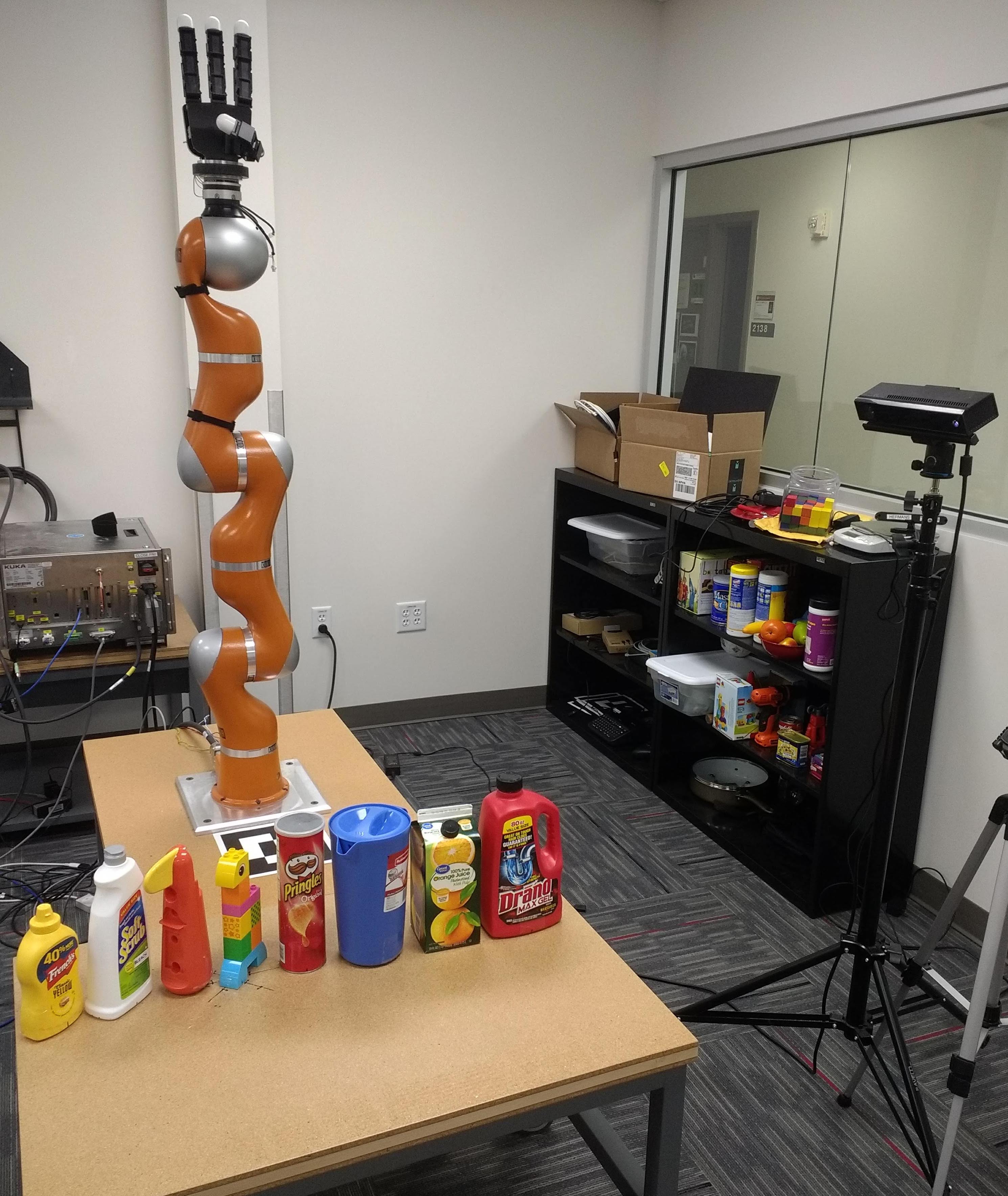}
  \caption{Our robot and camera setup for evaluating grasping in the real world with a KUKA LBR4 robot with Allegro end-effector. The objects are labeled from left to right as \emph{mustard}, \emph{soft-scrub}, \emph{airplane-toy}, \emph{Lego}, \emph{pringles}, \emph{pitcher}, \emph{juice-box}, and \emph{max-gel} respectively. }
  \label{fig:setup}
\end{figure}

We now combine our grasp success network with our reconstruction network for reconstruction-aware grasp synthesis, as described in Sec.~\ref{sec:grasp-synthesis}.

We implement our optimization in PAGMO~\cite{biscani2010global} and use SLSQP~\cite{kraft1989slsqp} to perform the optimization. Our analytic signed distance function (for the analytical environment collision constraint in Eq.~\ref{eq:const_collision}) is obtained using GJK~\cite{ong1997gilbert} and PQP~\cite{gottschalk1996obbtree}. We seed the optimization by sampling a grasp configuration from the grasp prior $g$ and use Inverse Kinematics (IK) to convert the 6DOF palm pose to a set of arm joints, constraining the IK solution to be above the table. We sample from the component of the prior that represents side grasps.

We complete our implementation of our formulation by using the PointSDF network for the robot-object collision constraint in Eq.~\ref{eq:const_learned_sdf} and using our grasp success network ``PointSDF-Fixed'' for the objective function in Eq.~\ref{eq:f_grasp_score}. We call this approach ``Reconstruction-Grasping.'' To provide a strong comparison, we introduce another approach that uses the ``PointSDF-Scratch'' model for the objective function and replaces the learned collision constraint in Eq.~\ref{eq:const_learned_sdf} with an analytical collision constraint. We provide this analytical collision model with the partial point cloud meshed at $32^3$ voxelization. This approach, which we call ``Partial-View-Grasping'', represents an approach that relies only on the partial geometric information available.

We test our grasp optimization with an Allegro hand mounted on a Kuka LBR4 arm. We use a Kinect 2 camera to get our point clouds, and place the camera such that the right-handed robot plans side-grasps near the occluded area of each object; this provides us with a geometrically difficult grasping problem. We use 6 objects from the YCB dataset~\cite{Calli2015} and two objects that were used by~\cite{Lu2019}. All objects as well as the full robot/camera setup are shown in Fig.~\ref{fig:setup}. Only the \emph{chips} object was part of the training dataset and all others are novel objects.

We place a single target object on the table and call the grasp optimization. In order to select good grasps, we accept the grasp plan if the likelihood of success from the learned grasp success network is above 0.6 \textit{and} if the combined grasp score with the prior (i.e., Eq.~\ref{eq:f_grasp_score}) is below 5.0. We allow up to five chances to satisfy this heuristic, each with a new seed from the prior; failure to find a sufficient plan given five chances is recorded as a failure. Given a successful robot joint configuration, the configuration is sent to MoveIt! to obtain a collision free trajectory to reach the grasp configuration. We allow up to three plans to be sent to MoveIt!, upon which continued lack of a motion plan is labeled a failure. For collision checking during motion planning, we provide the partial mesh of the object to MoveIt! for the ``Partial-View-Grasping'' method and the reconstructed mesh using PointSDF for the ``Reconstruction-Grasping'' method (evaluated to $128^3$ voxelization). Each successful MoveIt! generated trajectory is executed on the robot and the hand is closed with the same controller used during data collection. Finally, the palm is lifted to 15 cm. We chose a fixed test location on the table for the target object and rotated to 3 different angles, keeping the pose constant across the methods and objects. We also test the camera at two positions, offset by about 0.45 meters vertically (referred to as ``high'' and ``low'' scenarios respectively). This is to explore how robust each method is to various levels of occlusion: the lower setting yields heavier occlusion in the input point cloud.

\begin{figure}[]
  \centering
  \includegraphics[width=0.98\columnwidth]{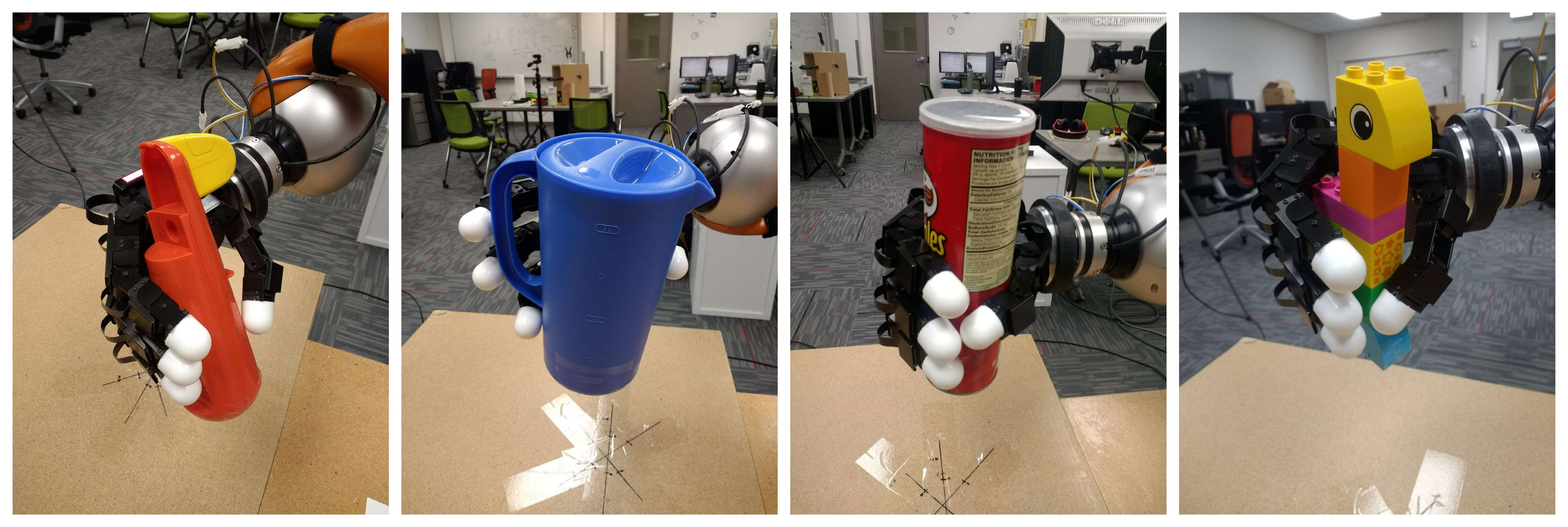}
  \caption{Representative grasps our approach executed on the physical robot.}
  \label{fig:real-robot-examples}
\end{figure}

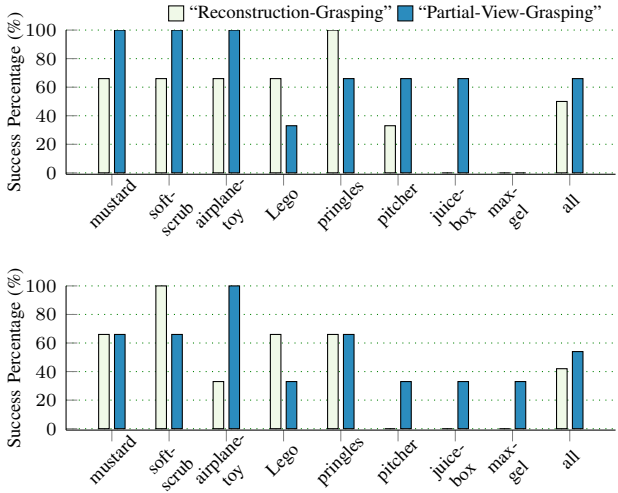
\begin{figure}[]
   \centering
   \begin{adjustbox}{minipage={0.98\columnwidth}}
    \begin{tikzpicture}
      \begin{axis}[
        ybar, ymin=-1, ymax=100,
        ylabel={Success Percentage (\%)},
        axis lines*=left,
        y label style={at={(axis description cs:0.045,0.5)},anchor=north},
        symbolic x coords={mustard, soft-scrub, airplane-toy, Lego, pringles, pitcher, juice-box, max-gel, all},
        xtick=data,
        x tick label style={font=\scriptsize,rotate=45,yshift=0.2cm,xshift=0.2cm,text width=0.8cm,align=center},
        xticklabels={mustard, soft-scrub, airplane-toy, Lego, pringles, pitcher, juice-box, max-gel, all},
        ymajorgrids=true,
        ticklabel style = {font=\scriptsize},
        legend style={font=\scriptsize, draw=none, fill=none},
        ylabel style = {font=\scriptsize},
        bar width = 4pt, height=3.5 cm, width=1.05\columnwidth,
        legend style={area legend, at={(1,1.25)}, anchor=north east, legend columns=3, },
        legend image code/.code={%
         \draw[#1] (0cm,-0.1cm) rectangle (2mm,1mm);},
       ]
        \addplot [fill=lightgreen] coordinates {
          (mustard, 66) (soft-scrub, 66) (airplane-toy, 66) (Lego, 66) (pringles, 100) (pitcher, 33) (juice-box, 0) (max-gel,0) (all,50)};

        \addplot [fill=darkblue] coordinates {
          (mustard, 100) (soft-scrub, 100) (airplane-toy, 100) (Lego, 33) (pringles, 66) (pitcher, 66) (juice-box, 66) (max-gel,0) (all, 66)};
        
        \legend{``Reconstruction-Grasping'', ``Partial-View-Grasping''}
      \end{axis}
    \end{tikzpicture}
    \begin{tikzpicture}
      \begin{axis}[
        ybar, ymin=-1, ymax=100,
        ylabel={Success Percentage (\%)},
        axis lines*=left,
        y label style={at={(axis description cs:0.045,0.5)},anchor=north},
        symbolic x coords={mustard, soft-scrub, airplane-toy, Lego, pringles, pitcher, juice-box, max-gel, all},
        xtick=data,
        x tick label style={font=\scriptsize,rotate=45,yshift=0.2cm,xshift=0.2cm,text width=0.8cm,align=center},
        xticklabels={mustard, soft-scrub, airplane-toy, Lego, pringles, pitcher, juice-box, max-gel, all},
  ymajorgrids=true,
        ticklabel style = {font=\scriptsize},
        ylabel style = {font=\scriptsize},
        bar width = 4pt, height=3.5 cm, width=1.05\columnwidth,
       ]
        \addplot [fill=lightgreen] coordinates {
          (mustard, 66) (soft-scrub, 100) (airplane-toy, 33) (Lego, 66) (pringles, 66) (pitcher, 0) (juice-box, 0) (max-gel, 0) (all, 42)};

        \addplot [fill=darkblue] coordinates {
          (mustard, 66) (soft-scrub, 66) (airplane-toy, 100) (Lego, 33) (pringles, 66) (pitcher, 33) (juice-box, 33) (max-gel,33) (all, 54)};
      \end{axis}
    \end{tikzpicture}
  \end{adjustbox}
    \caption{Success rates on real world grasp task. We plot the ``High'' camera setup on top and the ``Low'' camera setup (more occlusion) below.}
  \label{fig:plot_success}
\end{figure}

\begin{figure}
  \centering
  \includegraphics[width=\columnwidth]{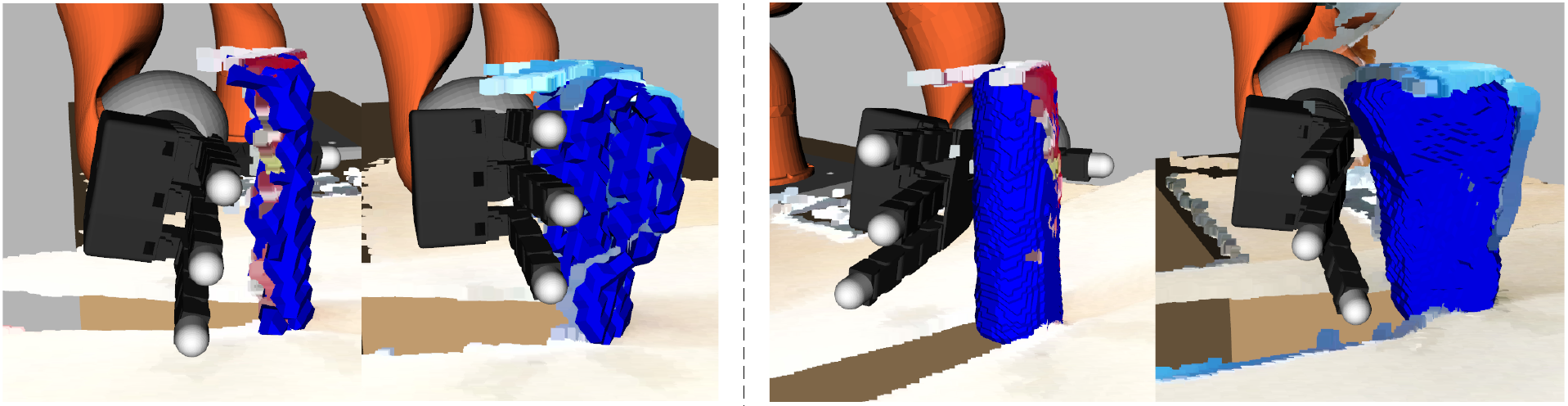}
  \caption{Qualitative examples of grasps where the ``Partial-View-Grasping'' (left) planned a grasp in direct collision with occluded space, while ``Reconstruction-Grasping'' (right) planned a geometry-aware grasp that avoids the occupied space.}
  \label{fig:occluded-grasps}
\end{figure}

We show some qualitative grasp execution examples from the ``Reconstruction-Grasping'' method in Fig.~\ref{fig:real-robot-examples}. We show the per-object and per-camera location performance for each approach in Fig.~\ref{fig:plot_success}. We ran a total of 96 grasps, split between the two approaches. Across both camera-setups, the ``Reconstruction-Grasping'' approach succeeded on 48\% of its grasp attempts (22 grasps). The ``Partial-View-Grasping'' approach succeeded on 60\% of its grasp attempts (29 grasps). The ``Reconstruction-Grasping'' approach has an average planning time of 37.81s as compared to ``Partial-View-Grasping'' approach's slightly shorter 36.55s.

From Fig.~\ref{fig:plot_success}, we see that the ``Reconstruction-Grasping'' approach struggles to grasp the larger objects from our object set (\emph{pitcher, juice-box, and max-gel}) and is roughly equivalent in performance across the other objects. We believe the lack of quantitative improvement is due, in part, to the increased difficulty the constraints over the full geometry impose on the optimization. Of the 26 failures, the ``Reconstruction-Grasping'' optimization failed to find a sufficient solution to our heuristic 11 times, as opposed to only 4 such failures for ``Partial-View-Grasping,'' indicating that the optimization was struggling to both fulfill the constraints and meet the heuristic cutoffs on grasp success likelihood.

In Fig.~\ref{fig:occluded-grasps} we show an example of a plan generated by the ``Partial-View-Grasping'' approach that yields a final grasp in contact with the target object. We see that the ``Reconstruction-Grasping'' approach can avoid this error through its understanding of the full geometry. Interestingly, we found actual grasp execution to be relatively robust to grasps in contact with the object; only one grasp failure for ``Partial-View-Grasping'' was clearly due to planning in contact with the object. Further work into determining when target geometry matters to the grasp success is needed to understand the extent to this robustness; intuitively, the qualitative results of Fig.~\ref{fig:occluded-grasps} could impact performance broadly, even if our results did not reflect this as clearly.



\section{Conclusion and Future Work}
\label{sec:conclusion}
We explore how geometry can be leveraged in grasp synthesis. We incorporate a learned signed distance function via a shared embedding space for grasp success prediction and add collision constraints to the grasp optimization to yield geometrically-aware grasp synthesis. Our results indicate that while our approach exhibits desirable qualitative geometric reasoning (e.g., Fig.~\ref{fig:occluded-grasps}), the difficulty of the optimization hurts our approach's quantitative gains. In future work, we hope to improve the constrained optimization efficacy and better understand when object geometry matters. We also plan to explore incorporating feedback from multiple viewpoints and tactile sensing~\cite{sundaralingam-icra2019-tactile-force-learning} to improve reconstruction and, in-turn, grasp success predictions. Finally, we hope to explore using the learned grasp success models as constraints for optimization of in-hand manipulation tasks~\cite{sundaralingam-icra2018-finger-gaiting}.




\section*{ACKNOWLEDGMENTS}
Mark Van der Merwe was supported in part by UROP at the University of Utah. Qingkai Lu and Balakumar Sundaralingam were supported by NSF Awards \#1846341 and \#1657596. The authors would like to thank M. Wilson, G. Tabor, and F. Ramos for helpful discussions. 

\bibliographystyle{IEEEtran}
\bibliography{main}  

\end{document}